\documentclass[conference]{IEEEtran}
\IEEEoverridecommandlockouts
\usepackage{cite}
\usepackage{amsmath,amssymb,amsfonts}
\usepackage{graphicx}
\usepackage{textcomp}
\usepackage{xcolor}
\usepackage{times}
\usepackage{latexsym}
\usepackage{amsmath}
\usepackage{multirow}
\usepackage{subfigure}
\usepackage{color}
\usepackage{graphicx}
\usepackage{tabu}
\usepackage{booktabs}
\def\BibTeX{{\rm B\kern-.05em{\sc i\kern-.025em b}\kern-.08em
    T\kern-.1667em\lower.7ex\hbox{E}\kern-.125emX}}
\begin{document}

\title{ReasonChainQA: Text-based Complex Question Answering with Explainable Evidence Chains}

\author{\IEEEauthorblockN{Minjun Zhu*\thanks{*National Laboratory of Pattern Recognition, Institute of Automation, CAS}\dag\thanks{\dag School of Artificial Intelligence, University of Chinese Academy of Sciences}}
\IEEEauthorblockA{
\textit{zhuminjun2020@ia.ac.cn}}
\and
\IEEEauthorblockN{Yixuan Weng*}
\IEEEauthorblockA{
\textit{wengsyx@gmail.com}}
\and
\IEEEauthorblockN{Shizhu He*\dag}
\IEEEauthorblockA{
\textit{shizhu.he@nlpr.ia.ac.cn}}
\and
\IEEEauthorblockN{Kang Liu*\dag}
\IEEEauthorblockA{
\textit{kliu@nlpr.ia.ac.cn}}
\and
\IEEEauthorblockN{Jun Zhao*\dag}
\IEEEauthorblockA{
\textit{jzhao@nlpr.ia.ac.cn}}
\and

}

\maketitle

\begin{abstract}
The ability of reasoning over evidence has received increasing attention in question answering (QA). Recently, natural language database (NLDB) conducts complex QA in knowledge base with textual evidences rather than structured representations, this task attracts a lot of attention because of the flexibility and richness of textual evidence. However, existing text-based complex question answering datasets fail to provide explicit reasoning process, while it's important for retrieval effectiveness and reasoning interpretability. Therefore, we present a benchmark \textbf{ReasonChainQA} with explanatory and explicit evidence chains. ReasonChainQA consists of two subtasks: answer generation and evidence chains extraction, it also contains higher diversity for multi-hop questions with varying depths, 12 reasoning types and 78 relations. To obtain high-quality textual evidences for answering complex question. Additional experiment on supervised and unsupervised retrieval fully indicates the significance of ReasonChainQA. Dataset and codes will be made publicly available upon accepted.
\end{abstract}


\section{Introduction}
Developing systems that can reason over explicit knowledge has attracted substantial attention in current AI research  \cite{clark2020transformers}. Complex Question Answering (\textbf{Complex QA}) tasks provide a comprehensive and quantitative way to measure these abilities, with evidence provided by structured knowledge bases (e.g.WikiData) \cite{lansurvey} or natural language texts (e.g. Wikipedia) \cite{zhu2021retrieving}. 
Considering the high cost of constructing structured knowledge bases, this paper focuses on complex QA over textual evidence. 

In textual complex QA tasks, prevailing datasets focus on multi-hop reasoning that requires the aggregation of evidence across multiple paragraphs to answer a question \cite{yang-etal-2018-hotpotqa}. In order to test the ability of reasoning over text in a more fine-grained way, WIKINLDB dataset \cite{JamesThorne2021DatabaseRO} constructing knowledge triples to sentence level evidence and introducing database queries into complex text QA task.
\begin{figure}[t]
	\centering

	\includegraphics[scale=0.43]{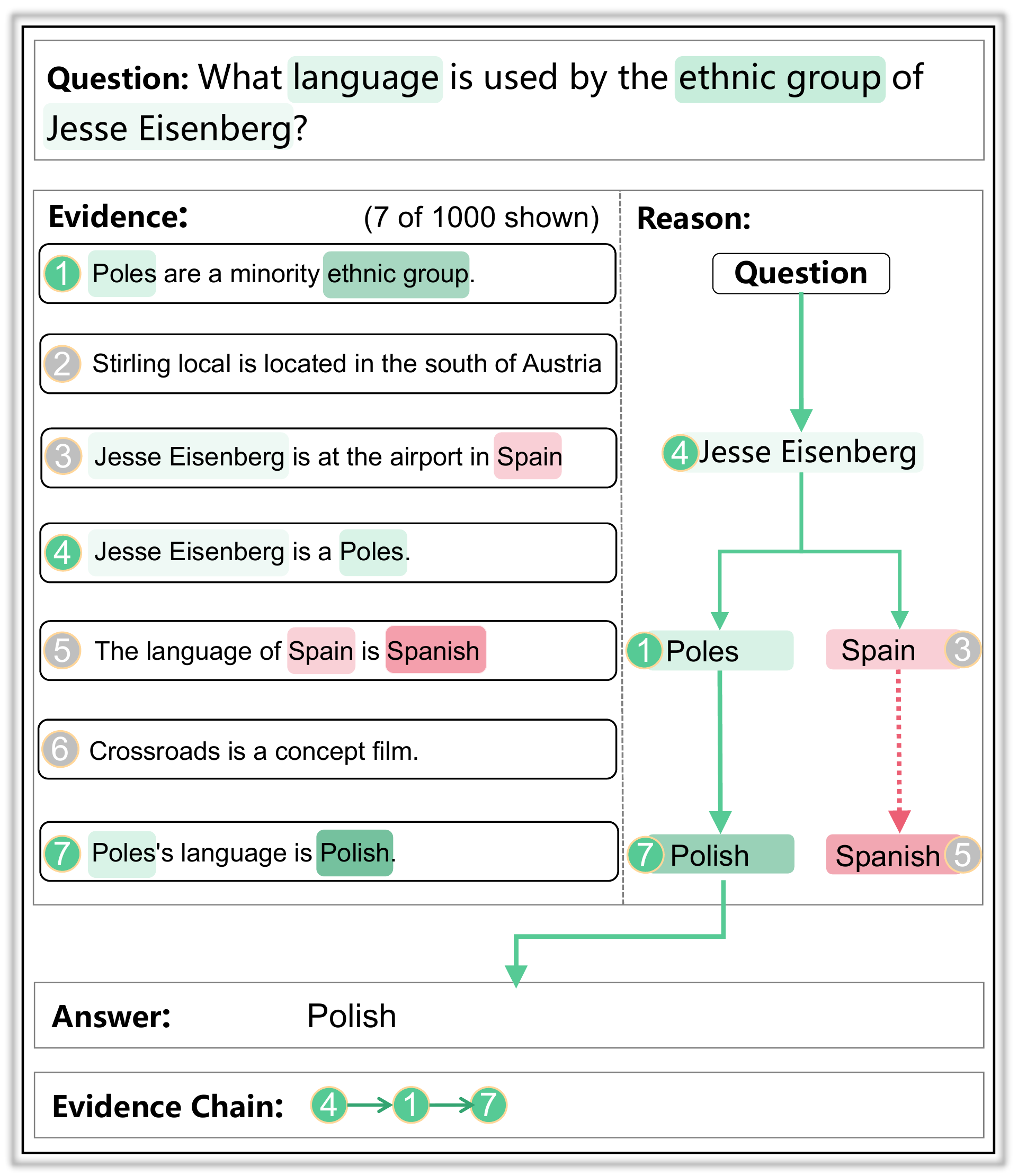}
		\caption{An example of the ReasonChainQA. Each sample contains a question, a set of textual evidence, a unique answer, and a chain of textual evidence with a complete path. }
	\label{text1}
	\vspace{-0.4cm}
\end{figure}
These datasets have sparked significant progress in textual complex QA. However, existing complex text QA datasets lack explicit, high-quality and in-depth reasoning process. Some datasets provide supporting passage sets without ordered evidence \cite{yang-etal-2018-hotpotqa,qi-etal-2021-answering} or organize evidence chains by overlap spans in passages \cite{WenhanXiong2020AnsweringCO}. This limitation will hinder the further advancement of reasoning over texts. 

In this work, we propose a step in this direction by introducing a new benchmark ReasonChainQA. Given a complex question based on textual evidence databases, models need to conduct two subtasks: (1) extracting evidence chains; (2) generating correct answer. questions. In ReasonChainQA dataset, knowledge facts were expressed as natural language sentences. An evidence chain is a logically connected sentence sequence from question to answer. % a fact relevant (or partially relevant) to give a reasonably supported
For example (as shown in figure \ref{text1}), the evidence chain of question ``\textit{What language is used by the ethnic group of Jesse Eisenbe?}'' is sentence $4\rightarrow1\rightarrow7$. Evidence chains provide systems with strong meaningful and explainable reasoning supervision about how the answer is derived.

In contrast to WIKINLDB dataset where queries are one-hop or two-hops with 4 reasoning types and 25 relations, and HotpotQA \cite{yang-etal-2018-hotpotqa}, where each query is answered over two passages with two reason types, ReasonChainQA not only expresses knowledge as NLDB, but also makes progress in questions' diversity and complexity. Specifically, our explicit reasoning chain depths vary from 1 to 7 (Table \ref{Table1}) with 12 inference types (Table \ref{table5}) and much more correlative relations. 

In summary, our contributions are as follows:

\textbf{1.} We present ReasonChainQA, the first textual database QA dataset with explicit evidence chains that covers diversified reason depths, relations and inference types. 
%  quantitatively analyze the performance of multi-hop retrieval model in the reasoning path.

\textbf{2.} We evaluated the performance of existing methods. It can more comprehensively reveal the prediction effect of different methods in the reasoning path, because our data set provides sequential annotation information

\textbf{3.} We conducted detailed experiments for reasonchain, providing a robust baseline method for subsequent research.

\begin{table*}[t]
		\centering \small
		\caption{Statistics of ReasonChainQA}
		\renewcommand\arraystretch{1.2} \setlength{\tabcolsep}{1.4mm}
		\begin{tabular}{c|ccc|c}
			\noalign{\hrule height 1pt}

			\textbf{Type} & Train &Dev &Test & Total \\
						\hline
			\textbf{Number} & 9505&1220 &1268 & 11993  \\
			\textbf{Avg. Question Token length} & 11.11 &11.28 &11.08 & 11.12 \\
			\textbf{Avg. Evidence Token length} & 9.88 &9.88 &9.85 & 9.87  \\
			\hline
			\textbf{Avg. Depth of Evidence Chain} & 2.13 &2.12 &2.13 & 2.13  \\
			\textbf{Depth of Evidence Chain $=$ 1} &3231 (33.99\%) &420 (34.43\%)&425 (33.52\%)&4076 (33.99\%)\\
			\textbf{Depth of Evidence Chain $=$ 2}& 2376 (25.00\%)&295 (24.18\%)& 336 (26.50\%)&3007 (25.07\%)\\
			\textbf{Depth of Evidence Chain $=$ 3} &3352 (35.27\%)&439 (35.98\%)& 424 (33.44\%)&4215 (35.15\%)\\
			\textbf{Depth of Evidence Chain $>=$ 4}& 546 (5.74\%)& 66 (5.41\%)&83 (6.54\%)&695 (5.79\%)\\
			\hline
			\noalign{\hrule height 1pt}
			
		\end{tabular}
		
		\label{Table1}
	\end{table*}

\section{Related Work}

\noindent\textbf{Complex Textual QA Tasks.}  
There is vast prior research on complex textual QA. Open-domain QA (OpenQA) tasks retrieve passages from large corpora \cite{EllenMVoorhees2000TheTQ,MatthewDunn2017SearchQAAN,joshi-etal-2017-triviaqa}. Complex QA  requires to deal well with passage structures, and infer answers from few paragraphs by deep reasoning. For example, DROP~\cite{dua-etal-2019-drop} needs to perform discrete numerical reasoning. HotpotQA \cite{yang-etal-2018-hotpotqa} covers multi-hop questions answered by collecting multiple documents. BeerQA \cite{qi-etal-2021-answering} integrates the datasets of different reasoning steps. RuleTaker \cite{clark2020transformers} reasons based on the rules. However, these tasks do not have evidence chains, and the impact of evidence chains on QA tasks has not been studied.

\noindent\textbf{Natural language databases} is a special textual QA task. It considers the questions similar to database queries, where system needs to reason and filter on a large number of textual evidence sentences. The bAbI \cite{JasonWeston2015TowardsAQ} needs to find the answer of one evidence from less than 20 evidences. WIKINLDB \cite{JamesThorne2021DatabaseRO} transforms knowledge triples to factual sentences, and answers can be discretely distributed in all corners of the database. These NLDBs represent knowledge without predefined schemas, and modular architectures on these datasets need to perform discrete reasoning over hundreds or thousands of textual evidence. ReasonChainQA provides explainable evidence chains after them, which can perform correct text reasoning.

\noindent\textbf{Complex Textual QA Methods.}
Many complex textual QA methods focus on the efficiency of documents retrieval \cite{karpukhin-etal-2020-dense} or the pseudo-tags of documents sequence modelling \cite{XinyuZhang2021AnswerCQ,AkariAsai2019LearningTR}. GRR \cite{AkariAsai2019LearningTR} recursively retrieves correct evidence in each step until the end-of-evidence token is generated. In previous textual QA tasks, retrieve-and-reading QA systems (open domain question answering) with pseudo chain tag indicate that a complete and sequential evidence path is vital in knowledge reasoning \cite{AkariAsai2019LearningTR,saha-etal-2020-prover,saha-etal-2021-multiprover}. However, most of these methods lack global evidence chain modeling, and their performance in long-chain is insufficient. Therefore, the introduction of reasoning chain with indefinite length can evaluate the performance of different methods more finely.

\section{Task Description}

\subsection{Problem Definition}
%ReasonChainQA refers to knowledge facts expressed as natural language sentences as database corpora. 
ReasonChainQA refers to natural language  sentences (textual evidence) as database, each sentence containing one or multiple knowledge facts. An example application of text databases is personal assistant.
%For text databases storing knowledge,
%such as daily experiences, friends and their preferences, 
%people will not organize all texts in predefined schema. They may ask the personal assistant to get the factual answers from the texts, like quering the database. 
For example, for answering the query \textit{``Among the feature films with a publication date after 2003, which one has the smallest duration?''}, QA models require set filter(e.g.,year, concept, duration), numeric comparison and facts' bridge over textual database.  
%\textit{"Hi~ Siri, where do I need to go on my trip tomorrow?" }Who is taller, Yao Ming or the spouse of Vanessa Laine Bryant?
% How many people does the Yao Ming's birth place have in 2016?
% TO solve these questions with The texts evidence database requires the agent to accurately perform reasoning tasks from various texts, so it is very necessary in scenarios such as personal assistants, 
% discrete 

For a question $Q_I$ in ReasonChainQA, it could be answered on a large-scale unsequenced factual texts database. To solve complex questions often requires specific logical reasoning path over multiple facts, we provide reasoning chains $E_I$, consisting of correct reasoning evidence $e_j^I$. 
%  we call the disordered evidence \mathbf{n}=
\begin{equation}
	E_I =  \bigcup_{j=1}^n \mathbf{e}^I_j
\end{equation}
For example, the evidence set of $E_I$ in Figure \ref{text1} is the evidence chain $R_I$=[1,4,7]. 
% We provide a question  that must be answered from .  And $I$ have a reasoning path $R_I$ constructed by correct evidence.
% supporting facts enable models to improve performance and make explainable predictions.

We propose a benchmark ReasonChainQA, and set two subtasks: (1) Learning to extract reasoning chains. (2) Generating corrects answers for complex questions. We believe that a more fine-grained evaluation of multi-hop question answering system can improve QA performance and make the prediction explainable.
%We find that reasoning chains can improve QA performance and make explainable prediction. 

\subsection{Challenge}

Mainstream systems developed with complex question answering over textual database tasks follow the retriever-and-reader architecture \cite{DanqiChen2017ReadingWT}, where the efficiency and effectiveness of evidence chain retrieval still remain to be great challenges.
 %In Complex Question answering over text database, prevailing systems follow retrieval-then-read architecture \cite{DanqiChen2017ReadingWT}, where efficiency and effectiveness of reason chains retrieval still remain to be great challenges.
 %In this subsection we discuss some significant challenges in ReasonchainQA, especially for knowledge evidence chain retrieval over text. % 补充一些文献和细节
 %ReasonChainQA follows the textual QA formulation, including retrieval and reading parts, 
%\item \textbf{Evidence Texts Representation.} % 长上下文本建模，知识三元组文本如何建模表示，和句子之间的关系
%Texts databases consist of thousands of fact sentences related to knowledge triples in semantic. 
%The ReasonChainQA sentences can be scattered in the databases, even though they are related closely in semantic and can form semantical graphs, which is different from traditional open domain question answering task (ODQA) based on separate long passages and documents. As a result, how to acquire reasoning-needed sentences (evidence), and model them with questions are the challenges in ResonChainQA.

\textbf{Retrieval Stopping Strategy.} % conditional retrieval 如何保证检索查准率和查全率。固定跳数和固定最大篇章不适合。
 To advance an efficient iterative retriever without dropping accuracy is still a challenge for evidence chain retrievers. 
 %Some retrievers iteratively get multiple supporting passages and evidence, but the retrieval efficiency would drop badly with the rising number of iterations, especially for complex queries with deep evidence chain. 
 Some existing work chooses to determine a fixed number of iterations \cite{WenhanXiong2020AnsweringCO} or a maximum number of retrieved passages \cite{YairFeldman2019MultiHopPR}, which can barely handle queries of varying depths in ReasonChainQA. Therefore, it's necessary to consider an iterative retrieval stopping strategy.
% To the best of our knowledge

\textbf{Evidence Chains Extraction.} % \textbf{beamsearch和MIPS不适合深度很深的链式文本检索方式。 
DPR \cite{karpukhin-etal-2020-dense}, MDR \cite{WenhanXiong2020AnsweringCO} and SSG \cite{JamesThorne2021DatabaseRO} use maximum inner-product search (MIPS) retrievers. GRR \cite{AkariAsai2019LearningTR} compares evidence with currently hidden state embedding independently, and fails to consider the context. These methods ignore the modelling of evidence chain in deep reasoning, which badly damages the overall performance of evidence chain retrieval. 
%which makes it impossible to retrieve deeper evidence and greatly reduces the retrieval accuracy of complex question and answer evidence base.
%GRR \cite{AkariAsai2019LearningTR} ends its retrieval only when the  is generated by its Recurrent Retriever.

\section{Evidence Chain Graph \label{ECG}}
We remodel each evidence chain as a directed acyclic graph, and filter the samples containing multiple parent nodes or multiple child nodes, and filter the samples containing multiple parent nodes or multiple child nodes. We redefined the relationship of each evidence node, including:
\textbf{Parent node} means that this node needs to be inferred from these nodes.

\textbf{Child node} means that these nodes can be obtained by reasoning based on this node.

\textbf{Sibling node} means that these two nodes are deduced by their parent nodes at the same time, which belongs to the parallel evidence relationship.

\textbf{No-cross node} means that in the process of reasoning, there is no relationship.

\begin{figure}[t]
	\centering
	\includegraphics[scale=0.23]{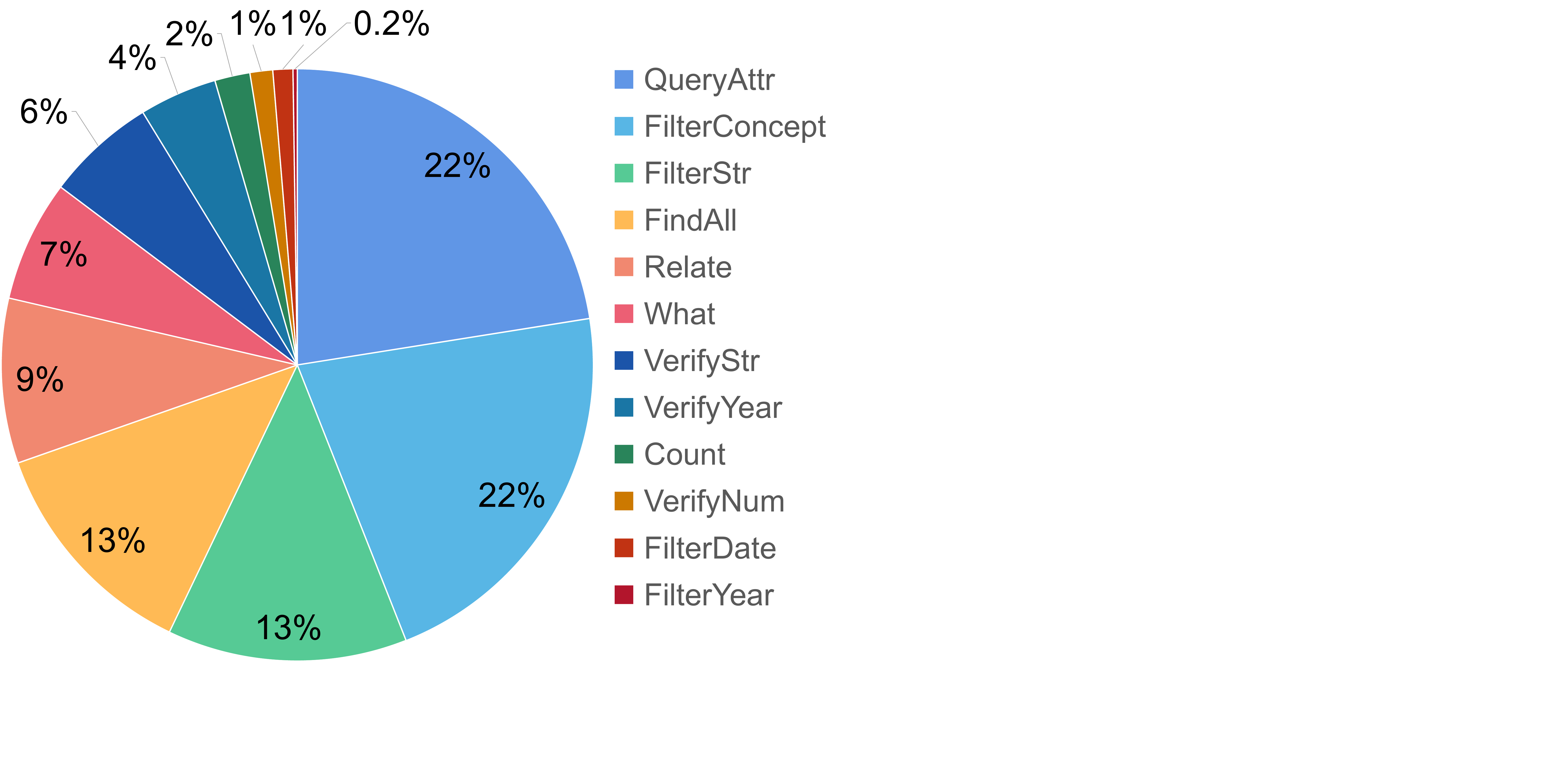}
	\caption{Percentage of each program in ReasonChainQA}
	\label{figure2}
	\vspace{-0.4cm}
\end{figure}

\begin{table*}[h]
	\centering  \small
	    \caption{Processing objective function  description for entity inference in ReasonChainQA (following KQA-PRO)}
   \begin{tabular}{|c|c|lc|}
   \hline
    \textbf{Program} & \textbf{Number}&\textbf{Example}    &\textbf{Example Output}      \\ \hline
    \textbf{QueryAttr}&8688& $\textbf{QueryAttr}(height)$ & $\textit{185cm} $  \\ 
    \textbf{FilterConcept}&8313& $\textbf{FilterConcept}(athlete)$ &  $\textit{Kobe Bryant} $ \\
\textbf{FilterStr}&5057& $\textbf{FilterStr}(gender, male)$ &  $\textit{(Entities, Facts)} $ \\ 
\textbf{FindAll}&4837& $\textbf{FindAll}()$ &  $\textit{All Entities} $\\
\textbf{Relate}&3476& $\textbf{Relate}(capital, forward)$ &  $\textit{Beijing} $ \\
\textbf{What}&2562& $\textbf{What}(Kobe Bryant)$ & $ \textit{Kobe Bryant} $\\
\textbf{VerifyStr}&2319&  $\textbf{VerifyStr}(male)$ & $\textit{True} $\\
\textbf{VerifyYear}&1641& $\textbf{VerifyYear}(1980, >)$ &$ \textit{False} $\\
\textbf{Count}&743& $\textbf{Count}(Entities)$ & $\textit{8} $\\
\textbf{VerifyNum}&479& $\textbf{VerifyNum}(20000 dollars, >)$ &$  \textit{False} $\\
\textbf{FilterDate}&421& $\textbf{FilterDate}(birthday, 1980/06/01, <)$ & $ \textit{(Entities, Facts)} $ \\
\textbf{FilterYear}&90& $\textbf{FilterYear}(birthday, 1980, =)$ &  $\textit{(Entities, Facts)} $\\ \hline

    \end{tabular}

	\label{table_query}
			\vspace{-0.2cm}
\end{table*}

\section{Dataset Collection}
%In this section we introduce the data collection of ReasonChainQA. 
In this section we introduce ReasonChainQA dataset, a diverse and explainable textual evidence-based complex question answering dataset with explicit reasoning chains. We first transform knowledge base triples from Wikidata \cite{DennyVrandecic2014WikidataAF} to natural language sentences, which providing us with abundant texts to get different size of databases. Then, to generate each example, we acquire texts evidence chains from structured triples according to query and program from KQA-Pro dataset \cite{KQA}. 
%Following setting in WIKINLDB, we provide datasets with evidence bases of different sizes {25,250,1000} in ReasonChainQA.

\subsection{Textural Evidence} %  
 We employ several methods to generate databases evidence grounded in Wikidata identifiers. 
 %We refer to the settings of KELM \cite{agarwal-etal-2021-knowledge} to convert the structured triple data of Wikidata into unstructured text. 
 For language diversity, we extend triples with same subject ones to generate more natural and informative facts as knowledge triple graphs. Then we train a text generation model by aligned corpus from KELM \cite{agarwal-etal-2021-knowledge}, which could generate a natural language evidence according to a knowledge triples graphs. 
 %,  searched from KB. the triples of the same subject and spliced them with templates, then input them into the mapping model. 
 For example, for a knowledge triple ``\{Winna, Lives in, London\}'', We find ``\{Winna, husband, Mike\}'' that related to Winna from the knowledge base. After combining them as a knowledge triple graph and inputing into the text generation model (based on T5), we can get ``\textit{Winna's husband is Mike and she lives in London}''. In this way, factual information of database evidence can be enriched. We also repeatedly utilize generation and post-processing methods to ensure faithfulness and diversity of evidence. Moreover. To avoid generation uncertainty \cite{DBLP:journals/corr/abs-2201-05273}, string matching is taken to ensure all triples appear in generated evidence. Data quality will be shown in V.C. Besides, we mainly collect irrelevant evidence from two aspects to enlarge database, facts with same subject and facts similar to queries. 
 % ReasonChainQA provide abundant text representation based on the sequence model \cite{2020t5,lewis-etal-2020-bart}.
% After that, the uncontrollability of the generated model \cite{ijcai2021-0612} will still bring the noise to the evidence text. We repeatedly generate the results by replacing the models with different weights. To ensure faithfulness, we use the string matching method to verify whether each piece of evidence completely contains all fact triple entities.

\subsection{QA and Evidence Chains} %

To get complex natural language questions and explicit reasoning process, we take over the questions and the corresponding answers in KQA Pro, which is a large-scale KBQA dataset with programs and SPARQLs. In order to  transform logical queries to evidence chains, firstly,  we select programs of KQA-Pro as the processing objective functions of fact reasoning, which contain 12 types. Secondly, we locate corresponding knowledge triples through the programs. Finally, we  redefine the relationship of each evidence node by filtering samples containing multiple parent nodes or multiple child nodes. The relationships between its different nodes include parent node, child node, sibling node and no-cross node. The details are shown in \ref{ECG}.  

\subsection{Analysis of ReasonChainQA\label{3.4}}
\textbf{Dataset Statistics.}
ReasonChainQA is a challenging large-scale OpenQA dataset composed of 11,993 QA pairs. The details are shown in Table \ref{Table1}. The reasoning length of the evidence chains is not fixed and the average length is 2.13. After that, to evaluate the generalization ability of the model on data sets of different scales, we construct three data sets of different sizes by randomly searching unrelated fact triples. Following setting in WIKINLDB, each of them has a database of 25, 250 and 1000 samples, respectively.

\noindent\textbf{Dataset Quality.}
%To measure the quality of generating factual texts from knowledge base triples, we randomly select 100 samples and manually evaluate factual accuracy and text fluency and complexity. 100/100 sentences can accurately map to according to triples, 27/100 sentences are mapped to multiple triples, and some sentences mapped to one triple are companied with qualifiers. 
To measure the quality of generating factual texts from knowledge base triples, we randomly select 100 samples and manually evaluate factual accuracy, text fluency and complexity. 100/100 sentences can accurately be mapped to according to triples, 27/100 sentences are mapped to multiple triples, and some sentences mapped to one triple are companied with qualifiers. 88/100 sentences can clearly understand the meaning. 99/100 of the samples can infer the correct answer.
\begin{table*}[h]
	\centering \small
	    \caption{Performance comparison of the variants of SOTA methods on ReasonChainQA dataset. We will use path sequence as a supervised comparison and not use path sequence as an unsupervised comparison. We highlight the best score in each column in \textbf{bold}, and the second best score with \underline{underline}.}
   \begin{tabular}{cc|ccc|ccc|ccc|c}
    \bottomrule
    \multicolumn{2}{c|}{\multirow{2}{*}{System}}                                                                                    & \multicolumn{3}{c|}{\textbf{25}}     & \multicolumn{3}{c|}{\textbf{250}}            & \multicolumn{3}{c|}{\textbf{1000}}  & \multirow{2}{*}{   Time$\downarrow$}    \\ 
    \multicolumn{2}{c|}{}                                                                                     &  EM$\uparrow$ &ED$\downarrow$ & F1$\uparrow$  &  EM$\uparrow  $&ED$\downarrow$ & F1$\uparrow$   &  EM$\uparrow$  &ED$\downarrow$ & F1$\uparrow$    &       \\  \midrule
    \multicolumn{2}{c|}{\textbf{Random}}                                                                       & 1.10\%&2.05 &8.24\% & 0.15\%&2.12&0.81\%&0.08\%&2.13&0.26\%&0.01X          \\ \midrule
    \multicolumn{1}{c|}{\multirow{3}{*}{\begin{tabular}[c]{@{}c@{}}Unsupervised\end{tabular}}} & \textbf{BM25} & 29.65\%&1.38&58.89\% &24.68\%  &1.52&47.38\% &22.32\% &1.57&  42.81\% & 0.03X          \\
    \multicolumn{1}{c|}{}                                                                       & \textbf{DPR}  &40.93\% & \underline{0.59} &66.77\%  & 34.78\%& \underline{0.65} & 61.94\%&25.32\% &\underline{0.75} & 60.05\% & 0.9X             \\
    \multicolumn{1}{c|}{}                                                                       & \textbf{SSG} &37.85\% &0.93 & 79.39\%&38.72\% &0.85 &31.91\% &37.38\% &0.84 & 11.10\%& 43X                   \\
    
     \midrule
    \multicolumn{1}{c|}{\multirow{2}{*}{\begin{tabular}[c]{@{}c@{}}Supervised\end{tabular}}}   & \textbf{GRR}  & \underline{50.55\%}&1.03&  \textbf{97.86\%}&\underline{46.77\%}& 1.15 &\textbf{87.66\%} &\underline{43.06\%}&1.27&\underline{81.37}\% & 1.1X              \\
    \multicolumn{1}{c|}{}                                                                       & \textbf{MDR} & \textbf{54.50\%} & \textbf{0.45} & \underline{85.43\%} &\textbf{51.74\%} & \textbf{0.48} & \underline{82.58\%}&\textbf{51.03\%}&\textbf{0.49}&\textbf{81.73\%}&3X \\
\bottomrule
    \end{tabular}

	\label{table1}

\end{table*}
\section{Experimental Results}
%Retrieval effectiveness and efficiency are both crucial factors for the deployment of an OpenQA system in practice, especially when it comes to the real-time scenarios.
\subsection{Setup}

In order to evaluate the performance of different methods in Reasonchain in a more comprehensive way, we divide them into two categories. The first one does not need to use sequence information as an unsupervised experiment, and the second one uses sequence information for supervision, which we regard as a supervised experiment.

In the unsupervised experiments, model is trained without the evidence chains' supervision, but evaluated with correct evidence chains. It is compared with some baseline systems without multi-hop retrieval. 

\subsection{Evaluation Metrics}
ReasonChainQA contians two subtasks, evidence chains extracting and complex question answering. The former subtask is the foundation for the latter, and can also provide interpretability for QA task. 

For sequential reasoning tasks, we divide them into unsupervised tasks and supervised tasks to evaluate the system. We use exact match and edit distance \cite{Levenshtein} to evaluate the effect of a system in multi-hop reasoning. Specifically, for unsupervised tasks, we do not provide relevant evidence chain labels but follow the setting of the original paper. For supervised tasks, we take our evidence chain tag as training, generate an evidence chain according to a model we require, and then calculate the matching degree between this evidence chain and the $R_I$ tag we provide. In addition, in order to more accurately evaluate the performance of the model on larger scale data, we provide additional test sets of $n = 250$ and $n = 1000$. For these, we compute recall (@k), which measures the fraction of times the correct document is found in the top-k predictions.

\begin{table}[t]
    \caption{Exact Match performance for QA tasks. We uniformly use the retrieved Topk evidence as input to the Reader for result generation.}
	\centering \small
   \begin{tabular}{cc|ccc}
   \hline
    \multicolumn{2}{c|}{\multirow{1}{*}{\textbf{QA  Exact Match}}}                                                                                 &   \multicolumn{1}{c}{\textbf{25}}    & \multicolumn{1}{c}{\textbf{250}}         &\multicolumn{1}{c}{\textbf{1000}}    \\ \hline
    \multicolumn{2}{c|}{\textbf{Random}} &33.44\% &29.57\% & 28.71\%               \\ \hline
    \multicolumn{1}{c|}{\multirow{3}{*}{\begin{tabular}[c]{@{}c@{}}Unsupervised\end{tabular}}} & \textbf{BM25} &57.10\% & 52.37\% & 50.95\%           \\
    \multicolumn{1}{c|}{}                                                                       & \textbf{DPR} &69.72\% &63.09\% &54.50\%                \\
    \multicolumn{1}{c|}{}                                                                       & \textbf{SSG} & 79.42\% & 64.51\% &  34.34\%                \\
 \midrule
    \multicolumn{1}{c|}{\multirow{2}{*}{\begin{tabular}[c]{@{}c@{}}Supervised\end{tabular}}}   & \textbf{GRR} &90.38\% &81.62\% &78.15\%               \\
    \multicolumn{1}{c|}{}                                                                       & \textbf{MDR} &70.27\% &68.45\% & 68.69\%\\
   \hline
    \end{tabular}

	\label{table2}

\end{table}

\subsection{Experimental Details}
We train the model using the PyTorch(1.8.2) \footnote{\url{https://pytorch.org}} \cite{NEURIPS2019_bdbca288} on the NVIDIA RTX3090 GPU and use the hugging-face\footnote{\url{https://github.com/huggingface/transformers}} \cite{wolf-etal-2020-transformers} framework. For all methods, the bert-base-uncased \footnote{\url{https://huggingface.co/bert-base-uncased}} model are chosen for feature extraction. The pretrained contextual encoders are of base size with 12 layers. We input the output vector of BERT into the average pooling layer to obtain the text vector. We use the AdamW \cite{IlyaLoshchilov2018DecoupledWD} as the optimizer with the warm-up \cite{7780459}, and fine tune the whole model with a learning rate of 1e-5. A bidirectional single-layer LSTM is randomly initialized as the Decoder, hidden state is 384, to obtain a new $V_H^{t+1}$ with dimension 768 from input [$V_Q$, $V_E^t$, $V_H^t$]. We set the maximum length of 40, delete the excess. We use linear decay of learning rate and gradient clipping of 1e-6. The dropout \cite{NitishSrivastava2014DropoutAS} of 0.1 is applied to prevent overfitting. When calculating the prediction time, Random and BM25 are performed on the CPU and the rest of the systems uniformly use the GPU RTX3090 for prediction time statistics.

 To ensure fairness, all hyperparameters are adjusted in the dev set. In all our experiments, at the end of each epoch of training, we will test in the Dev dataset, and select the highest model (Mainly depends on F1) to predict in the Test dataset. We report the results in the Test dataset. All the experimental results are repeated three times, and the highest and lowest scores are removed.

\begin{table}[h]
	\centering \small
	\caption{ Hyper-parameter settings.}
   \begin{tabular}{c|c}
   \hline
    \textbf{Hyper-parameter} & \textbf{Value}         \\ \hline
    Encoder Hidden Size & 768 \\
    Dropout & 0.1\\
    Learning Rate & 1e-5\\
    Batch size & 24\\
    Num Epochs & 10\\
    Beam & 5 \\
    \hline

    \end{tabular}
    
	\label{table5}
\end{table}

\subsection{Performance of Benchmarks}
Following previous work \cite{AkariAsai2019LearningTR,JamesThorne2021DatabaseRO}, we report Exact Match (EM), Edit Distance (ED), and F1 in Table \ref{table1}. We can observe that: Firstly, when the evidence chain information of the training set \cite{AkariAsai2019LearningTR} is limited as an unsupervised situation. Almost all methods cannot show excellent performance.

Secondly, for all methods, with the improvement of chain level accuracy, the accuracy of retrieval is also improving, which proves that improving the reasoning performance of the evidence chain can further assist retrieval.

\subsection{The Effectiveness of QA}
We follow the standard Retriever-Reader architecture \cite{DanqiChen2017ReadingWT} and evaluate the effectiveness of different searchers in QA tasks through fixed readers. In Table \ref{table2}, we employ T5 models fine-tuned on golden evidence as a Reader. We uniformly use the same fine-tuned T5 \cite{2020t5} model with a learning rate of 1e-4 for the final question and answer accuracy. It as the reader for fairness consistency and report the EM scores of reader results from different systems.

The retrieval evidence chains or sets from different systems are input to the reader. For the unsupervised system, we uniformly extract the results from Top-k retrieval evidence where k = length of evidence chains. For the supervised system, we take the retrieval model for generation (this means that k is judged by the model), after which the generated results are used as input to the reader model. We train five readers with different random numbers and take the average score as the reported result.

We can find that a more orderly reasoning chain can help readers better understand the structure of evidence and further find the correct answer.

\section{Conclusions}
In this study, we propose a diverse and explainable complex question answering database with explicit reasoning chain. ReasonChainQA provides accurate path retrieval information to help the system learn interpretable evidential reasoning.  We evaluate six baseline methods on three benchmark datasets at different scales. In the experiment, we evaluated in detail the modeling ability of different methods of evidence chain sequence, which was lacking in previous experiments. We further showed the importance of the evidence chain to question and answer tasks.

\bibliography{custom}
\bibliographystyle{IEEEtran}
\end{document}